\documentclass{article}
\usepackage{spconf,amsmath,graphicx}

\usepackage{times}
\usepackage{epsfig}
\usepackage{graphicx}
\usepackage{amsmath}
\usepackage{amssymb}
\usepackage{subfig}
\usepackage{booktabs}
\usepackage{multirow}      
\usepackage{multicol}   
\usepackage{placeins}
\usepackage{enumitem}

\title{Compressing Deep CNNs using Basis Representation and Spectral Fine-tuning}
\name{Muhammad Tayab\textsuperscript{1,2} , Fahad Ahmad Khan\textsuperscript{3}, Abhijit Mahalanobis\textsuperscript{1,2}}
\address{Center for Research in Computer Vision\textsuperscript{1}\\
Department of Computer Science\textsuperscript{2}\\
University of Central Florida\textsuperscript{3}, USA}
\begin{document}
%
\maketitle
\begin{abstract}
We propose an efficient and straightforward method for compressing deep convolutional neural networks (CNNs) that uses basis filters to represent the convolutional layers, and optimizes the performance of the compressed network directly in the basis space. Specifically, any spatial convolution layer of the CNN can be replaced by two successive convolution layers: the first is a set of three-dimensional orthonormal basis filters, followed by a layer of one-dimensional filters that represents the original spatial filters in the basis space. We jointly fine-tune both the basis and the filter representation to directly mitigate any performance loss due to the truncation. Generality of the proposed approach is demonstrated by applying it to several well known deep CNN architectures and data sets for image classification and object detection. We also present the execution time and power usage at different compression levels on the Xavier Jetson AGX processor. 

\end{abstract}

\section{Introduction}
\label{sec:intro}
\begin{figure}
\begin{center}
 {\includegraphics[clip, trim=6.7cm 0cm 6.5cm 0cm, scale=0.4] {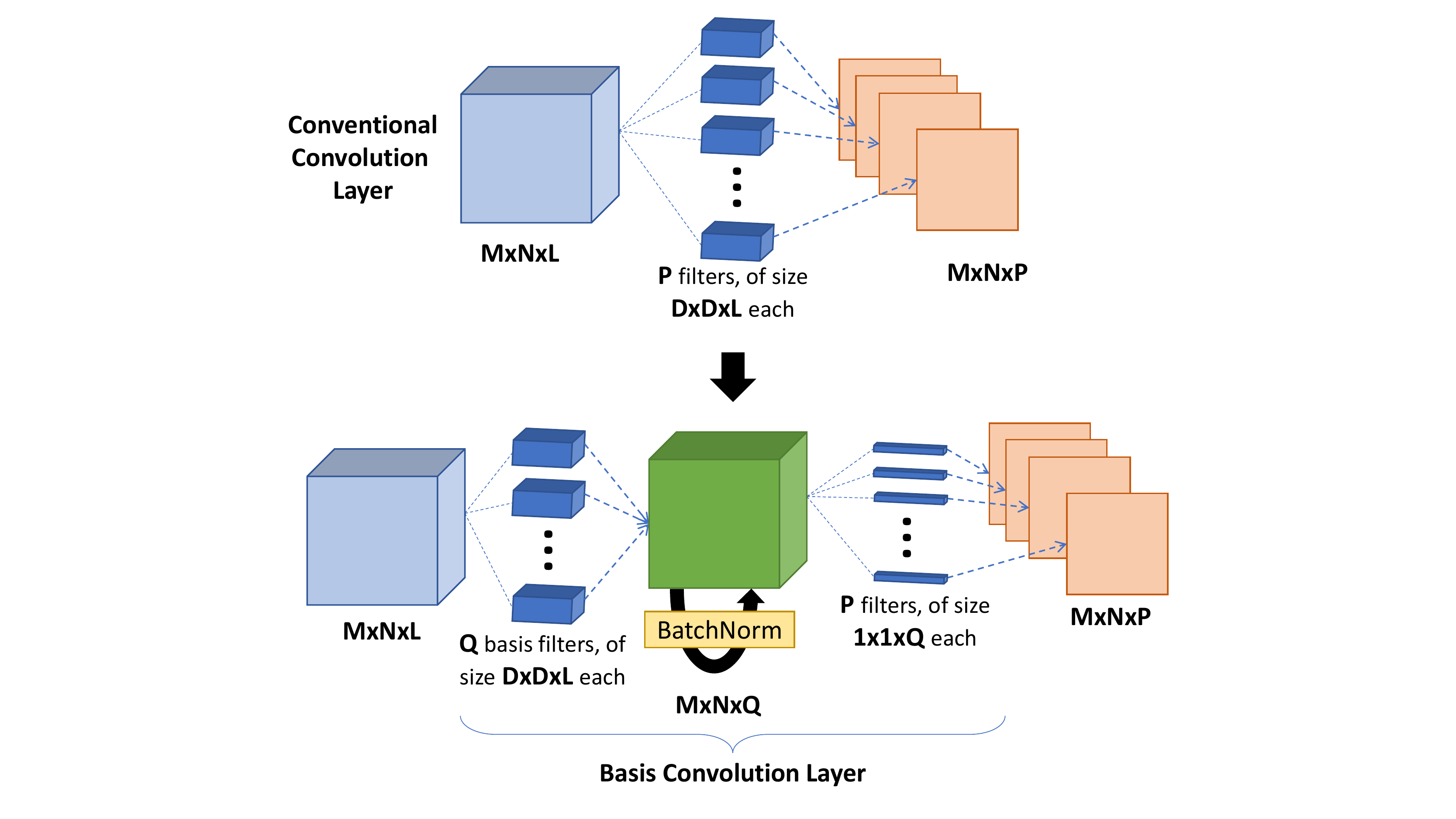}}
\end{center}
 \caption{The network is compressed by replacing each convolution layer (top) in a CNN with the basis convolution layer (bottom). Q basis filters are initialized with the top ranked eigen vectors of P original filters, with Q $<$ P. }
 \label{fig:basisconv}
\end{figure}

\begin{figure*}
\begin{center}
  \includegraphics[clip, trim=0.5cm 8.4cm 0.5cm 3.5cm, scale=0.45, width=\textwidth]{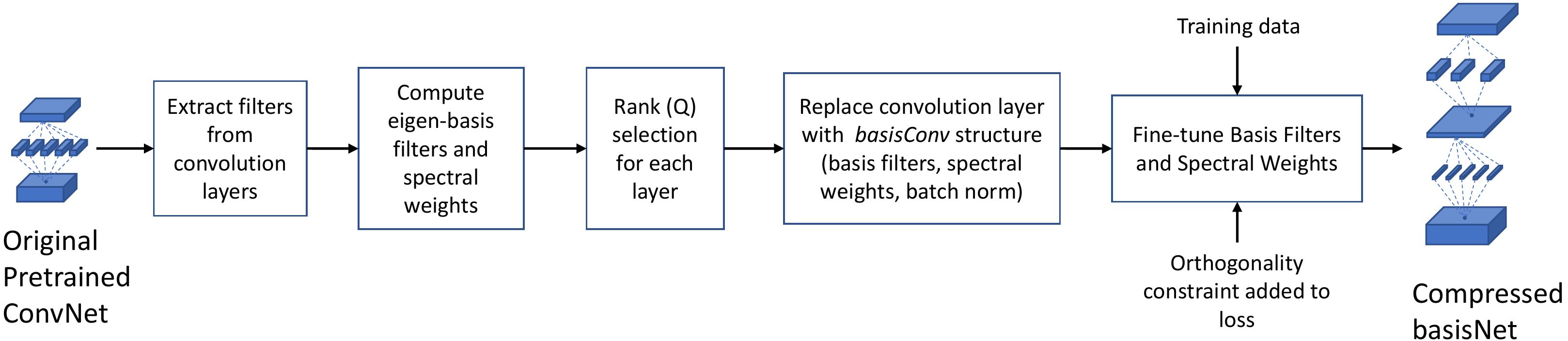}
\end{center}
  \caption{Schematic diagram of our compression framework.}
  \label{fig:recipe}
\end{figure*}

While there has been a tremendous surge in convolutional neural networks and their applications in computer vision, relatively little is still understood about how information is learned and stored in the network. This is evidenced by the fact that researchers have successfully proposed different approaches for compressing a network after it has been trained, including techniques like pruning weights \cite{optimalbrain, Chen2015CompressingNN}, assuming row-column separability and applying low rank approximations for computational gains \cite{Denton2014ExploitingLS-denton, 9-Xu2020TRPTR} and using basis representation
\cite{Jaderberg2014SpeedingUC-jaderberg, 3-Alvarez2017CompressionawareTO, 4-Li2018ConstrainedOB, 6-Minnehan2019CascadedPE, 9-Xu2020TRPTR}
to approximate the filter kernels.  Although the results are impressive,  existing compression techniques are not always easy to implement using standard deep learning tool boxes. In this paper, we are motivated by the observation that the original filters can be represented as weighted linear combinations of a set of 3D \textit{basis filters}  with one-dimensional \textit{weight} kernels as shown in Figure \ref{fig:basisconv}. While compression is achieved by using fewer basis filters, we show that these basis filters can be jointly  \textit{finetuned} along with the weight kernels  to compensate for any loss in performance due to truncation, and to thereby achieve state of the art results. 
The representation of spatial filters as a linear combination of orthogonal basis filters is also known as \textit{spectral decomposition}, and the weights comprise the corresponding \textit{spectra} of the filters. 
Since our approach updates the basis filters and their weighted contribution to the overall result, we refer to this process as \textit{Spectral Fine Tuning} (SFT). Not only does SFT reduce the number of learnable parameters, but also the weights are statistically uncorrelated, and therefore adapt much faster than the conventional finetuning of spatial domain filters. Figure \ref{fig:recipe} shows the main steps in our proposed approach for network compression.

Low rank decomposition has been extensively used for neural network compression in the past, but our work is differs from previously published papers in the manner in which the filters are represented and finetuned using basis filters and weights. For example methods such as \cite{3-Alvarez2017CompressionawareTO} enforce filter rank reduction while training the network, thereby making it more compressible than its normally trained counterpart. In contrast, \cite{9-Xu2020TRPTR} achieves compression by combining channel wise low rank approximation and with separable one-dimensional spatial filters. In \cite{4-Li2018ConstrainedOB} multiply-accumulate operations are reduced via constrained optimization. 
Low rank factorization and pruning are combined in \cite{6-Minnehan2019CascadedPE} by cascading the low rank projections of filters in the current layer to the next layer. In comparison, our method is relatively straight forward (as depicted in Figure \ref{fig:recipe}), and yet provides competitive results. 

\section{Basis Representation and Learning }
\label{sec:Basis-Representation-and-Learning}
Consider the fundamental convolution operation in any given layer of a convolutional neural network. Assume that an input block of data $x(m,n,l)$ (such as the activations or output of the previous layer) is convolved with a set of 3D filters $h_k (m,n,l), \ k=1$\dots$P$. The output $y_k (m,n)$ can be expressed as $y_k(m, n) = x(m, n, l) \ * \ h_k(m, n, l), \qquad 1 \leq k \leq p$
where $*$ represents the convolution operation. The filters can be expressed as a linear combination of Q basis functions $f_i (m,n,l), \quad i=1$\dots$Q$, such that $h_k (m, n, l) = \sum_{i=1}^{Q} w_{k} (i) \cdot f_i(m, n, l) $
where $w_{k} (i)$ are the weights of linear combination. Using this representation, the output can be expressed as 
\begin{equation}
\label{eq:3}
y'_k(m, n) = \sum_{i=1}^{Q} w_{k} (i) \cdot [x(m, n, l) * f_i(m, n, l)], \quad 1 \leq k \leq P   
\end{equation}
The key observation is that the Q convolution terms  $z_i (m,n)=x(m,n,l)*f_i (m,n,l)$ need to be computed only once, and they are common to all $P$ outputs  $y'_k (m,n)$.  These can be stacked together to form the 3D intermediate result $z(m,n,i)$ while the weights can be treated as $1 \times 1\times Q$ filter $w_k (i)$. Therefore, the outputs $y'_k (m,n)$ are simply the convolution of two, i.e 
\begin{equation}
y'_k (m,n)=w_k (i)*z(m,n,i)
\end{equation}
We refer to this construct using two successive convolutions as \textit{BasisConv}.

\textbf{Compression of pretrained Networks and Spectral Fine Tuning: }SFT is the simultaneous learning of $w_k (i)$ and $f_i (m,n,l)$ in Equation \ref{eq:3} (across all layers of the network) to mitigate overall performance loss due the choice of $Q < P$. We now discuss how the basis filters $f_i (m,n,l)$ and spectral weights $w_k (i)$ are initialized, and the orthogonality criteria introduced in the loss function to ensure that the basis representation condition is preserved.

It is well known that eigen decomposition results in a compact basis that minimizes the reconstruction error achieved by a linear combination of basis functions. We therefore initially choose $f_i (m,n,l)$ as the eigen filters that represent the sub-space in which the original filters $h_k (m,n,l)$ lie. The method for obtaining these is also well-known and straightforward. Basically, we define the $LD^2 \times 1$ dimensional vector  $\mathbf{h_k}$ as a vectorized representation of  $h_k (m,n,l)$, and construct the matrix $\mathbf{A}= [\mathbf{h_1} \ \mathbf{h_2} \ ... \ \mathbf{h_P}]$ with $\mathbf{h_k}$ as its columns. The eigenvectors of $\mathbf{A A^T}$ represent the sub-space of the filters, and satisfy the relation $\mathbf{A A^T f_i}=\lambda_i \mathbf{f_i}$, where $\mathbf{f_i}$ are the eigenvectors, and $\lambda_i$ are the corresponding eigenvalues. 
The eigen filter  $f_i (m,n,l)$ is readily obtained by re-ordering the elements of the eigenvector $\mathbf{f_i}$ into a $D \times D \times L$ array. It should be noted that these matrix-vector manipulations do not alter the inherent tensor structure and relations between the individual elements of the 3D filters. We select a small subset of eigenvectors which correspond to the largest Q eigen-values that best represent the dominant coordinates of the filters’ subspace using the metric $t = \frac{ \sum_{i=1}^{Q} \lambda_i}   { \sum_{i=1}^{LD^2} \lambda_i}$ to choose Q such that most of the relevant information is retained in the selected eigen-vectors. Using this criteria, $Q$ is chosen for each layer such that it results in no more than 3\% drop in overall performance.\par
The decomposition of the filters $h_k$ of any given layer of the network can be succinctly expressed in matrix vector notation by defining $\mathbf{F}= [\mathbf{f_1} \ \mathbf{f_2} \ ... \ \mathbf{f_Q}]$  (i.e. the matrix of eigenvectors of the filters for that layer)  so that $ \mathbf{h_k} = \mathbf{Fw_k}$
and $\mathbf{w_k} = [w_{k} (1)\ w_{k} (2)\ ... \ w_{k} (Q)]^T$ is a $Q \times 1$ vector of weights. Since $\mathbf{F^T F=I}$ (i.e. the identity matrix), the weights are easily obtained by computing $\mathbf{w_k} = \mathbf{F^T h_k}$.

SFT converges much faster than conventional finetuning of the original spatial filters. This is partly due to the fact that the elements of the weight vectors $\mathbf{w_k}$ are statistically uncorrelated. This is easy to show by noting that $\mathbf{A}= \mathbf{F}[\mathbf{w_1} \ \mathbf{w_2} \ ... \ \mathbf{w_P}]=\mathbf{FW}$
where $\mathbf{W}=[\mathbf{w_1} \ \mathbf{w_2} \ ... \ \mathbf{w_P}]$ is a matrix whose columns are the weight vectors associated with the filters in the corresponding columns of $\mathbf{A}$. Therefore, $\mathbf{A A^T}=\mathbf{FWW^TF^T}$.
It is also true that $\mathbf{A A^T}=\mathbf{F{\Delta}F^T}$
where $\mathbf{\Delta}$ is a diagonal matrix with the eigenvalues $\lambda_i$ as its diagonal elements. Hence, $\mathbf{WW}^T=\mathbf{\Delta}$ is a diagonal matrix and $\sum_{k=1}^{P} {w_k (i)}^2=\lambda_i , \text{   and   }   \sum_{k=1}^{P} {w_k (i)}{w_k (j)}= 0$. Therefore the convergences of each element of $\mathbf{w_k}$ is statistically uncorrelated with the behavior of the other elements, which implies that the corresponding eigen vector also contributes to the overall learning process in an uncorrelated manner.
Thus far, we have discussed fine-tuning the weights while holding the basis filters fixed as eigen-vectors of the original filters. However refining the basis filters can improve performance of the compressed network. To finetune the basis filters, we must ensure that they remain an orthonormal set (so that the weights continue to be a spectral representation in the new basis). This is achieved by including the following term in the overall cost function 
\begin{equation}
\mathbf{J}_f = \frac{\alpha}{{Q}}  \sum_{i=1}^Q (1-{\mathbf{f}_i}^T \mathbf{f}_i)^2 + \frac{2(1-\alpha)}{Q(Q-1)} \sum_{i=1}^Q \sum_{j=i+1}^Q ({\mathbf{f}_i}^T\mathbf{f}_j)^2
\label{euq:norm}
\end{equation}

where $\mathbf{f}_i$ are the individual basis filters, $Q$ is the number of such filters in a given layer, and $a$ is a positive number between 1 and 0. The first term in  $\mathbf{J}_f$ causes the $L2$ norm of $\mathbf{f}_i , i=1 ... Q$ to be as close to unity as possible, while the second term ensures the filters remain orthogonal. The parameter $a$ can be set to 0.5 to equally emphasize both term, or selected as necessary to balance the trade-off between the two terms. Finally, $f_i (m,n,l)$ and  ${w_k}(i)$ are both jointly fine-tuned to minimize the sum of $\mathbf{J}_f$ and the cost function (such as cross-entropy loss) used for training the original network.

\begin{table}[t]
\centering
\small
{\renewcommand{\arraystretch}{1}
\begin{tabular}{c | c c c } 
\hline
\textbf{Method} & \textbf{3$\times$} & \textbf{4$\times$} & \textbf{5$\times$} \\ 
\hline \hline
Jaderberg et. al.\cite{Jaderberg2014SpeedingUC-jaderberg}  & 87.6 & 80.2 & 60.2 \\
Zhang et. al.\cite{Zhang2016-AcceleratingVD} (asym.) & 89.5 & 89.0 & 87.9 \\
Zhang et. al.\cite{Zhang2016-AcceleratingVD} (asym. fine-tuned) & 89.9 & 89.6 & 88.9  \\
CP (fine-tuned) \cite{He2017ChannelPF-CP}  & - & 88.9 & 88.2 \\
ELR \cite{Wang2018-ELR}  & - & 89.4 & 88.8 \\
Li et. al. \cite{Li2016PruningFF-Lietal}  & - & 81.3 & 67.9 \\
Li et. al. \cite{4-Li2018ConstrainedOB}  & 88.9 & - & - \\
Minnehan et. al. \cite{6-Minnehan2019CascadedPE}  & - & - & 89.4 \\
\hline
\textit{Ours} & \textbf{90.6} & \textbf{89.6} & \textbf{89.4}  \\ 
\hline
\end{tabular}
}
\caption{Top-5 accuracy of VGG16 pretrained on ImageNet and compressed to 3x, 4x and 5x speedup ratios.}
\label{tab:pami15-imagenet}
\end{table}

\begin{table*}[t]
\centering
\small
{\renewcommand{\arraystretch}{1}
\begin{tabular}{c || c c c | c c c | c c c | c c c}
\hline
\multirow{2}{3em}{\textbf{Model}} & \multicolumn{3}{c}{\textbf{FLOPs (Billions)}} & \multicolumn{3}{|c}{\textbf{Params (Millions)}} & \multicolumn{3}{|c|}{\textbf{\# of Filters}} & \multicolumn{3}{c}{\textbf{Accuracy (\%)}} \\
 & CNN & BasisNet & $\downarrow$\% & CNN & BasisNet & $\downarrow$\% & CNN & BasisNet & $\downarrow$\% & CNN & BasisNet & $\downarrow$  \\
\hline 
Alexnet & 0.03 & 0.01 & 76.20       & 2.47 & 0.68 & 72.33   & 1152 & 256 & 77.78 & 43.87 & 44.68 & -0.81 \\
VGG16 & 0.63 & 0.12 & 80.44     & 14.71 & 1.67 & 88.68  & 4224 & 501 & 88.14 & 68.72 & 69.53 & -0.81 \\ 
Resnet110 & 0.51 & 0.16 & 68.65     & 1.73 & 0.63 & 63.41   & 4144 & 1271 & 69.33 & 71.99 & 71.67 & 0.32 \\ 
Densenet190 & 18.61 & 3.44 & 81.52  & 25.60 & 6.98 & 72.73  & 20290 & 4116 & 79.71 & 82.83 & 82.28 & 0.55 \\ 
\hline
\end{tabular}
}
\caption{ Results of compressing four different networks on CIFAR100 shows significant FLOPs and parameter reduction is achieved with minimal loss of accuracy. [$\downarrow$\%] denotes percent decrease in the metric compared to the baseline model.}
\label{tab:cifar100}
\end{table*}

\section{Experiments}
\label{sec:Experiments}
\begin{table}[t]
\centering
\small
\setlength{\tabcolsep}{6pt}
{\renewcommand{\arraystretch}{1}
\begin{tabular}{c|cccc}
\hline
\multirow{2}{4em}{\textbf{Method}} & \textbf{  FLOPs  } & \textbf{Params. } & \multicolumn{2}{c}{\textbf{Accuracy (\%)}} \\
  & (\%$\downarrow$) & (\%$\downarrow$) & Top1 & Top5 \\
\hline
\hline
SFP \cite{SFP}                          & 41.80 &   -   & 74.61 & 92.06 \\
IMP \cite{molchanov2019taylor-IMP}      & 45.00 & 51.48 & 74.50 & - \\
CP \cite{He2017ChannelPF-CP}            & 50.00 & -     & - & 90.80 \\
LFC \cite{LFC}                          & 50.00 & -     & 73.40 & 91.40 \\
ELR \cite{Wang2018-ELR}                 & 50.00 & -     & - & 91.20 \\
GDP \cite{Lin2018AcceleratingCN-GDP}    & 51.30 & -     & 71.89 & 90.719 \\
FPGM \cite{He2018FilterPV-FPGM}         & 53.50 & -     & 74.83 & 92.32 \\
DCP \cite{Zhuang2018-DCPA}              & 55.76 & 51.45 & 74.95 & 92.32 \\
GBN \cite{zhonghui2019-decorator}       & 55.06 & 53.40 & 75.18 & 92.41 \\
LFPC \cite{he2020learning-LFPC}         & 60.80 & -     & 74.46 & 92.04 \\
HRank \cite{lin2020-HRank}              & 62.10 & 46.13 & 71.98 & 91.01 \\
Alvarez et. al. \cite{3-Alvarez2017CompressionawareTO}  & - & 27.0 & 75.2 & 91.01 \\
TRP1+Nu \cite{9-Xu2020TRPTR}  & - & 44.0 & 74.06 & 92.07 \\
\hline
\textit{Ours (A)}       & \text{51.28} & \text{49.28} & \textbf{75.85} & \textbf{92.84}\\
\textit{Ours (B)}       & \textbf{62.68} & \textbf{61.16} & \text{74.96} & \text{92.44}\\
\hline
\end{tabular}
}
\caption{Results for Resnet-50 pretrained on ImageNet. Our compression configuration (A) achieves higest Top-1 and Top-5 accuracy at comparable level of compression.}
\label{tab:resnet50-imagenet}
\end{table}

\begin{table}[t]
\centering
\small
{\renewcommand{\arraystretch}{1}
\begin{tabular}{c | c c c  } 
\hline
\textbf{Method} & \textbf{FLOPs $\downarrow$\%} & \textbf{Params $\downarrow$\%} & \textbf{Acc. \%}  \\ 
\hline \hline
Li et. al. \cite{Li2016PruningFF-Lietal} & 27.6 & 13.7 & 93.06 \\
NISP \cite{Yu2017-NISP} & 43.6 & 42.6 & 93.09  \\ 
DCP-A \cite{Zhuang2018-DCPA} & 47.1 & \textbf{70.3} & 93.10  \\ 
CP \cite{He2017ChannelPF-CP} & 50.0 & - & 91.80 \\ 
AMC \cite{He2018-AMC} & 50.0 & - & 91.90 \\ 
GBN \cite{zhonghui2019-decorator} & 60.1 & 53.5 & \textbf{93.43} \\ 
LFPC \cite{he2020learning-LFPC} & 52.9 & - & 93.24 \\ 
FPGM \cite{He2018FilterPV-FPGM} & 52.6 & - & 93.26 \\ 
Minnehan et. al \cite{6-Minnehan2019CascadedPE} & 48.8 & - & 93.22 \\
TRP1 \cite{9-Xu2020TRPTR} & 56.7 & - & 92.77 \\
\hline
\textit{Ours} & \textbf{64.1}  & 60.2  &  93.01 \\
\hline
\end{tabular}
}
\caption{Results for Resent-56 on CIFAR10. Our compressed model achieves highest compression while attaining comparable or in some cases better accuracy.}
\label{tab:resnet56-cifar10}
\end{table}


\textbf{Data Sets and CNN Architectures:}
We performed our experiments on various publicly available datasets. 
These include image classification datasets 
\textit{CIFAR10}, \textit{CIFAR100}, \textit{ImageNet} and object detection dataset \textit{MS-COCO}. 
\par

\textbf{VGG16 on ImageNet:} We tested our method on VGG16 pretrained on ImageNet by compressing the model to 3x, 4x and 5x speedup ratios. Results of this experiment are presented in Table \ref{tab:pami15-imagenet}. Compared with other methods our basis VGG16 achieves better Top-5 accuracy on all speedup ratios.

\textbf{CIFAR-100 with Different Networks:} Table \ref{tab:cifar100} shows the results of compressing four different networks using the proposed method, and their performance on the CIFAR-100 data set. For instance, compressing DenseNet190 reduces FLOPS by 81\%, and the number of parameters by 73\% with a minor loss of 0.55\% in performance accuracy. Similar gains are also noted for the other architectures as well. 

\textbf{Resnet50 on ImageNet:} We also tested our method on Resnet with 50 convolution layers, pretrained on ImageNet by compressing the model at two different levels of compression: 51.28\% and 61.68\%. Results of this experiment and comparisons with other methods are presented in Table \ref{tab:resnet50-imagenet}.
When compared with existing state of art methods like IMP\cite{molchanov2019taylor-IMP}, CP\cite{He2017ChannelPF-CP}, FPGM\cite{He2018FilterPV-FPGM}, GBN\cite{zhonghui2019-decorator}, LFPC\cite{he2020learning-LFPC} and HRank\cite{lin2020-HRank} our model (A) achieves better accuracy for the similar levels of compression. 
Our second compression configuration (B) achieves best compression both in terms of FLOPs and trainable parameters, while keeping Top-1 and Top-5 accuracy comparable to other methods.

\textbf{Resnet56 on CIFAR10:} For CIFAR10 dataset we tested our method on Resnet56 model and compared it with other similar methods. Comparative Results are shown in Table \ref{tab:resnet56-cifar10}. 





\textbf{Results on Object Detection: }To demonstrate our method's effectiveness irrespective of the model architecture, dataset and the objective function, we tested it on object detection problem. We used 2 predefined variants of 
YOLOv3 \cite{Redmon2018YOLOv3AI-yolov3} object detector pretrained on MS-COCO dataset. These variants are YOLOv3-tiny \cite{Redmon2018YOLOv3AI-yolov3}, which contains 13 convolution layers and a much deeper model YOLOv3-416 \cite{Redmon2018YOLOv3AI-yolov3}, which contains 75 convolution layers. 
For YOLOv3-tiny our method was able to achieve 51.26\% reduction in FLOPs and 67.73\% reduction in model parameters with 29.6 MAP while the MAP for baseline model is 32.9. We also compressed YOLO-416 to 2x and 3x speedup ratios. These compressed models attain 53.83 and 55.34 MAP score respectively, as compared to the baseline model which has 55.40 MAP.

\begin{table}[t]
\centering
\small
{\renewcommand{\arraystretch}{1}
\begin{tabular}{c | c c c c c } 
\hline
\textbf{Method} & \textbf{Baseline} & \textbf{2$\times$} & \textbf{3$\times$} & \textbf{4$\times$} & \textbf{5$\times$} \\ 
\hline \hline
Inference time (ms)  & 40   & 30   & 26   & 25   & 23 \\
Power (W)            & 6.01 & 5.55 & 5.15 & 5.08 & 4.94 \\
Energy (Joule)       & 0.24 & 0.17 & 0.14 & 0.12 & 0.11 \\
Top5 Accuracy (\%)   & 90.4 & - & 90.6 & 89.6 & 89.4\\
\hline
\end{tabular}
}
\caption{VGG16 inference time, power and energy consumption on NVIDIA Jetson AGX Xavier, with various compression factors of 2x, 3x, 4x and 5x}
\label{tab:processor_speed}
\end{table}


 \textbf{Implementation On Edge Processors:} We also measured the GPU speed up of our method on Nvidia Jetson Nano by compressing the VGG16 to reduce the FLOPs by 50\% and 66\%. Our method achieves 35.6\% and 45.3\% speedup respectively. Table \ref{tab:processor_speed} shows the inference time, power and energy consumption of VGG16 on the NVIDIA Jetson AGX Xavier (trained on ImageNet) using compression factors of 2x, 3x, 4x and 5x. We see that total energy use can be reduced in half with less than 1\% loss in accuracy. Thus, we verified that the compressed model not only runs faster, but also requires less energy when implemented in edge processors that operate with limited resources such as GPU memory and power. It should be noted that we did not implement our method in C++ but used the python based Pytorch instead. We believe that an optimized C++ cuda implementation can achieve even better performance on GPUs. 
\section{Summary and Discussions}
We have presented a straightforward yet effective method for compressing deep CNNs that outperforms most other state of the art techniques in terms of reduction in FLOPS and number of parameters with negligible performance loss. Despite the overwhelming interest in neural network compression in recent years, there is still a lack of standardized definitions for metrics such as FLOPs.  Some methods choose to compress the network to arbitrary reduction in FLOPs and report the model's accuracy, while others compress the network in such a way that accuracy of the resulting model is within a certain range of base model. The inference time of the compressed model is an important metric that should be considered, but introduces its own complexities due to dependencies on libraries and hardware platform. We also note that power utilization is an important consideration for edge processors, and should be included as a metric for comparing compressed networks. 
\bibliographystyle{IEEEbib}
\bibliography{ICIP2021}

\begin{thebibliography}{10}

\bibitem{optimalbrain}
Yann~Le Cun, John~S. Denker, and Sara~A. Solla,
\newblock ``Optimal brain damage,''
\newblock in {\em Advances in Neural Information Processing Systems}. 1990, pp.
  598--605, Morgan Kaufmann.

\bibitem{Chen2015CompressingNN}
Wenlin Chen, James~T. Wilson, Stephen Tyree, Kilian~Q. Weinberger, and Yixin
  Chen,
\newblock ``Compressing neural networks with the hashing trick,''
\newblock in {\em ICML}, 2015.

\bibitem{Denton2014ExploitingLS-denton}
Emily~L. Denton, Wojciech Zaremba, Joan Bruna, Yann LeCun, and Rob Fergus,
\newblock ``Exploiting linear structure within convolutional networks for
  efficient evaluation,''
\newblock in {\em NIPS}, 2014.

\bibitem{9-Xu2020TRPTR}
Yuhui Xu, Yuxi Li, Shuai Zhang, Wei Wen, B.~Wang, Yingyong Qi, Yiran Chen,
  W.~Lin, and Hongkai Xiong,
\newblock ``Trp: Trained rank pruning for efficient deep neural networks,''
\newblock in {\em IJCAI}, 2020.

\bibitem{Jaderberg2014SpeedingUC-jaderberg}
Max Jaderberg, Andrea Vedaldi, and Andrew Zisserman,
\newblock ``Speeding up convolutional neural networks with low rank
  expansions,''
\newblock {\em CoRR}, vol. abs/1405.3866, 2014.

\bibitem{3-Alvarez2017CompressionawareTO}
Jose~M. Alvarez and M.~Salzmann,
\newblock ``Compression-aware training of deep networks,''
\newblock in {\em NIPS}, 2017.

\bibitem{4-Li2018ConstrainedOB}
Chong Li and C.~R. Shi,
\newblock ``Constrained optimization based low-rank approximation of deep
  neural networks,''
\newblock in {\em ECCV}, 2018.

\bibitem{6-Minnehan2019CascadedPE}
Breton Minnehan and A.~Savakis,
\newblock ``Cascaded projection: End-to-end network compression and
  acceleration,''
\newblock {\em 2019 IEEE/CVF Conference on Computer Vision and Pattern
  Recognition (CVPR)}, pp. 10707--10716, 2019.

\bibitem{Zhang2016-AcceleratingVD}
Xiangyu Zhang, Jianhua Zou, Kaiming He, and Jian Sun,
\newblock ``Accelerating very deep convolutional networks for classification
  and detection,''
\newblock {\em IEEE Transactions on Pattern Analysis and Machine Intelligence},
  vol. 38, pp. 1943--1955, 2016.

\bibitem{He2017ChannelPF-CP}
Yihui He, Xiangyu Zhang, and Jian Sun,
\newblock ``Channel pruning for accelerating very deep neural networks,''
\newblock {\em 2017 IEEE International Conference on Computer Vision (ICCV)},
  pp. 1398--1406, 2017.

\bibitem{Wang2018-ELR}
Dong Wang, L.~Zhou, Xueni Zhang, Xiao Bai, and J.~Zhou,
\newblock ``Exploring linear relationship in feature map subspace for convnets
  compression,''
\newblock {\em ArXiv}, vol. abs/1803.05729, 2018.

\bibitem{Li2016PruningFF-Lietal}
Hao Li, Asim Kadav, Igor Durdanovic, Hanan Samet, and Hans~Peter Graf,
\newblock ``Pruning filters for efficient convnets,''
\newblock in {\em International Conference on Learning Representations}, 2016.

\bibitem{SFP}
Yang He, Guoliang Kang, Xuanyi Dong, Yanwei Fu, and Yi~Yang,
\newblock ``Soft filter pruning for accelerating deep convolutional neural
  networks,''
\newblock in {\em Proceedings of the 27th International Joint Conference on
  Artificial Intelligence}. 2018, IJCAI’18, p. 2234–2240, AAAI Press.

\bibitem{molchanov2019taylor-IMP}
Pavlo Molchanov, Arun Mallya, Stephen Tyree, Iuri Frosio, and Jan Kautz,
\newblock ``Importance estimation for neural network pruning,''
\newblock in {\em Proceedings of the IEEE Conference on Computer Vision and
  Pattern Recognition}, 2019.

\bibitem{LFC}
Pravendra Singh, {Vinay Kumar} Verma, Piyush Rai, and {Vinay P.} Namboodiri,
\newblock ``Leveraging filter correlations for deep model compression,''
\newblock in {\em IEEE Winter Conference on Applications of Computer Vision,
  WACV 2020}. pp. 824--833, IEEE.

\bibitem{Lin2018AcceleratingCN-GDP}
Shaohui Lin, Rongrong Ji, Yuchao Li, Yongjian Wu, Feiyue Huang, and Baochang
  Zhang,
\newblock ``Accelerating convolutional networks via global \& dynamic filter
  pruning,''
\newblock in {\em IJCAI}, 2018.

\bibitem{He2018FilterPV-FPGM}
Yang He, Ping Liu, Ziwei Wang, Zhilan Hu, and Yang Yang,
\newblock ``Filter pruning via geometric median for deep convolutional neural
  networks acceleration,''
\newblock {\em 2019 IEEE/CVF Conference on Computer Vision and Pattern
  Recognition (CVPR)}, pp. 4335--4344, 2018.

\bibitem{Zhuang2018-DCPA}
Zhuangwei Zhuang, Mingkui Tan, Bohan Zhuang, Jing Liu, Yong Guo, Qingyao Wu,
  Junzhou Huang, and Jin-Hui Zhu,
\newblock ``Discrimination-aware channel pruning for deep neural networks,''
\newblock in {\em NeurIPS}, 2018.

\bibitem{zhonghui2019-decorator}
Zhonghui You, Kun Yan, Jinmian Ye, Meng Ma, and Ping Wang,
\newblock ``Gate decorator: Global filter pruning method for accelerating deep
  convolutional neural networks,''
\newblock in {\em Advances in Neural Information Processing Systems (NeurIPS)},
  2019.

\bibitem{he2020learning-LFPC}
Yang He, Yuhang Ding, Ping Liu, Linchao Zhu, Hanwang Zhang, and Yi~Yang,
\newblock ``Learning filter pruning criteria for deep convolutional neural
  networks acceleration,''
\newblock in {\em Proceedings of the IEEE Conference on Computer Vision and
  Pattern Recognition (CVPR)}, 2020.

\bibitem{lin2020-HRank}
Mingbao Lin, Rongrong Ji, Yan Wang, Yichen Zhang, Baochang Zhang, Yonghong
  Tian, and Ling Shao,
\newblock ``Hrank: Filter pruning using high-rank feature map,''
\newblock in {\em Proceedings of the IEEE/CVF Conference on Computer Vision and
  Pattern Recognition (CVPR)}, 2020, pp. 1529--1538.

\bibitem{Yu2017-NISP}
Ruichi Yu, Ang Li, Chun-Fu Chen, Jui-Hsin Lai, Vlad~I. Morariu, Xintong Han,
  Mingfei Gao, Ching-Yung Lin, and Larry~S. Davis,
\newblock ``Nisp: Pruning networks using neuron importance score propagation,''
\newblock {\em 2018 IEEE/CVF Conference on Computer Vision and Pattern
  Recognition}, pp. 9194--9203, 2017.

\bibitem{He2018-AMC}
Yihui He, Ji~Lin, Zhijian Liu, Hanrui Wang, Li-Jia Li, and Song Han,
\newblock ``Amc: Automl for model compression and acceleration on mobile
  devices,''
\newblock in {\em ECCV}, 2018.

\bibitem{Redmon2018YOLOv3AI-yolov3}
Joseph Redmon and Ali Farhadi,
\newblock ``Yolov3: An incremental improvement,''
\newblock {\em ArXiv}, vol. abs/1804.02767, 2018.

\end{thebibliography}

\end{document}